\documentclass{article}

\PassOptionsToPackage{numbers, sort&compress}{natbib}
\usepackage[final]{nips_2016}

\usepackage[utf8]{inputenc} 
\usepackage[T1]{fontenc}    
\usepackage{url}            
\usepackage{booktabs}       
\usepackage{amsfonts}       
\usepackage{amsmath}
\usepackage{amssymb}
\usepackage{nicefrac}       
\usepackage{microtype}      
\usepackage{subcaption}
\usepackage[normalem]{ulem}

\usepackage{todonotes}

\usepackage{graphicx}

\newcommand*\diff{\mathop{}\!\mathrm{d}}

\usepackage{setspace}

\title{Probabilistic structure discovery in time series data}
\author{
  David Janz \\
  University of Cambridge \\
  \texttt{dj343@cam.ac.uk} \\
  \And{}
  Brooks Paige \\
  University of Oxford  \\
  \texttt{brooks@robots.ox.ac.uk} \\
  \And{}
  Tom Rainforth \\
  University of Oxford \\
  \texttt{twgr@robots.ox.ac.uk} \\
  \And{}
  Jan-Willem van de Meent \\
  Northeastern University \\
  \texttt{j.vandemeent@northeastern.edu} \\
  \And{}
  Frank Wood \\
  University of Oxford  \\
  \texttt{fwood@robots.ox.ac.uk}
}

\begin{document}

\maketitle

\vspace{-5pt}

\begin{abstract}
	\vspace{-5pt}
Existing methods for structure discovery in time series data
construct interpretable, compositional kernels for Gaussian process regression models.
While the learned Gaussian process model provides posterior mean and variance estimates,
typically the structure is learned via a greedy optimization procedure. This restricts the space of possible solutions and leads to over-confident uncertainty estimates.
We introduce a fully Bayesian approach, inferring a full posterior over structures, which more reliably captures the uncertainty of the model.
\end{abstract}

\vspace{-5pt}

\section{Introduction}
\vspace{-5pt}
How much of the data science process can we automate?  Many techniques such as random forests \cite{breiman2001random}, neural networks \cite{lecun2015deep} and support vector machines \cite{cortes1995support} can deliver outstanding predictive performance when used in an out-of-the-box fashion \cite{fernandez2014we,rainforth2015canonical}, while the rise of so-called automated machine learning \cite{feurer2015efficient} is providing increased automation of the full data-science pipeline.  However, such methods tend to focus only on predictive performance, giving crude estimates for uncertainty and providing negligible human insight into the data generation process.  Such approaches are not, in isolation, sufficient in the strife for automated data science.

This paper focuses on automatic methods for model construction and selection in the context of models built up compositionally from simple atomic parts \cite{grosse2014model,duvenaud2013structure}.
These models can be expressive and powerful while still maintaining human interpretability, a key characteristic for automatic ML systems that are to be trusted by end users. Specifically, our method builds upon automatic Bayesian covariance discovery \cite{lloyd2014automatic} (ABCD), which employs search over the kernel structure and parameters of a Gaussian process \cite{rasmussen2006gaussian} (GP) for regressing time series data.

We suppose that some time series data
$\mathcal{D} = \{(y_n,x_n) : n=1,\dots,N \}$ was generated by a latent function $f$ with additive Gaussian noise. We consider a GP prior $f \sim \mathcal{GP}(0, k_{\theta})$ for some positive semi-definite kernel function
$k_{\theta}(x, x')$
parametrised by a vector of hyperparameters $\theta = \left(\theta_1,\dots, \theta_P \right)^{\top}$.
Noting that the set of positive semi-definite functions $\mathbf{S^+}$ has the property $z, z' \in \mathbf{S^+}$ implies $z+z' \in \mathbf{S^+}$ and $z\times z' \in \mathbf{S^+}$ \cite{rasmussen2006gaussian},
we define a space of kernels as the space of hierarchical products and sums of base kernels $b \in \mathcal{B}$. Here $\mathcal{B}$ consists of the linear ($\textsc{Lin}$), squared exponential ($\textsc{SE}$), periodic ($\textsc{Per}$) and rational quadratic ($\textsc{RQ}$) kernels.
ABCD selects a single $k_{\theta}$ using a greedy optimisation based search that maximises the marginal likelihood of the model.

We expand on this by integrating out $k_\theta$.  In particular, we explore inferring the distribution over possible kernel compositions, to provide better a better uncertainty estimate and allow exploration of the multiple modes in the model space. 
Integrating out $k_\theta$ has previously been attempted by \citet{schaechtle2015probabilistic} in a restricted setting, where set of possible kernel combinations is finite and heavily limited;
we extend this to use a true stochastic context-free grammar as a prior on kernel structure. 
Our approach is implemented within the probabilistic programming framework of Anglican \cite{wood-aistats-2014}, allowing integration with other systems in a probabilistic modelling pipeline.

\begin{figure}
\centering
\begin{subfigure}[b]{0.47\textwidth}
  \includegraphics[width=\textwidth]{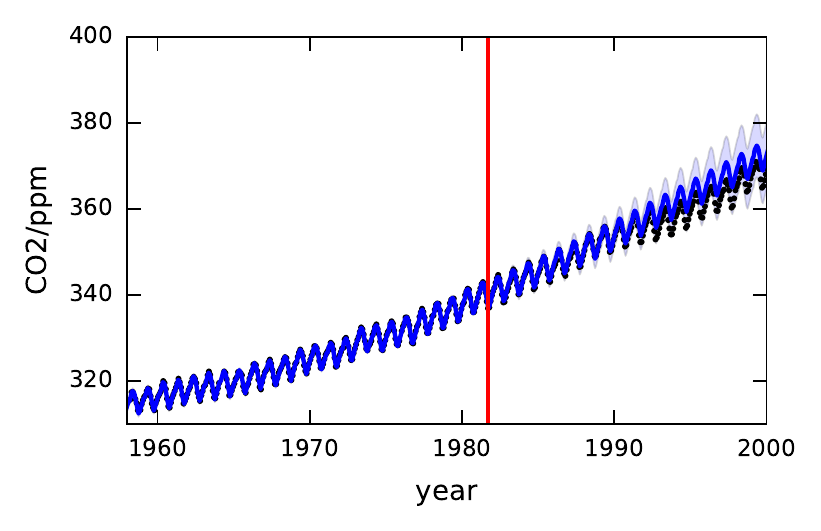}
\vspace{-1.5em}
  \caption{Optimisation}
  \label{subfig:opt}
\end{subfigure}
\hspace{0.04\textwidth}
\begin{subfigure}[b]{0.47\textwidth}
  \includegraphics[width=\textwidth]{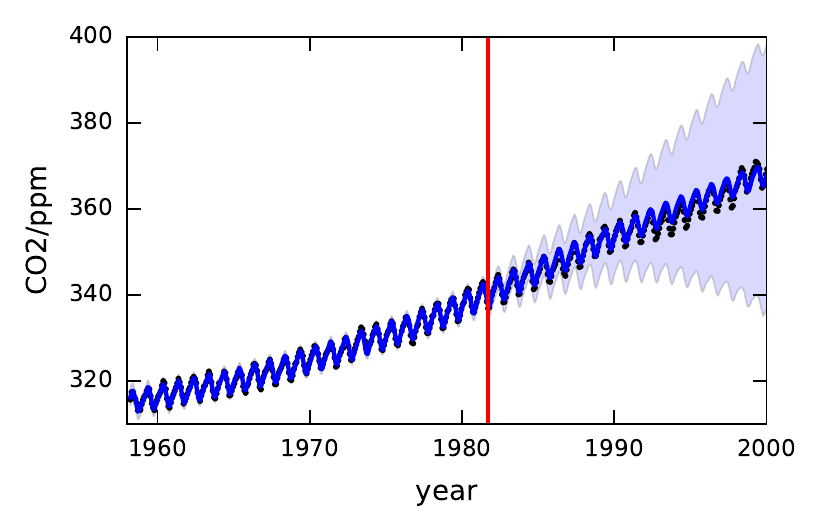}
\vspace{-1.5em}
  \caption{Inference}
  \label{subfig:inf}
\end{subfigure}
\caption{Regression of the Maunu Lao CO$_{2}$ dataset \cite{rasmussen2006gaussian}.
Models are trained on data from 1957 through 1983 (red line); we compare the extrapolation to collected data from 1984 through 2000. Median plotted, 10-90 percentile region shaded.
(\ref{subfig:opt}) The learned best-fit model and model parameters using ABCD fits the training data closely, but yields overconfident predictions in the extrapolation.
(\ref{subfig:inf}) Averaging over multiple plausible models yields better-calibrated uncertainty estimates.}
\label{fig:main_result}
\end{figure}

\begin{figure}
\centering
\captionsetup[subfigure]{labelformat=empty}
\begin{subfigure}[b]{0.325\textwidth}
  \includegraphics[width=\textwidth]{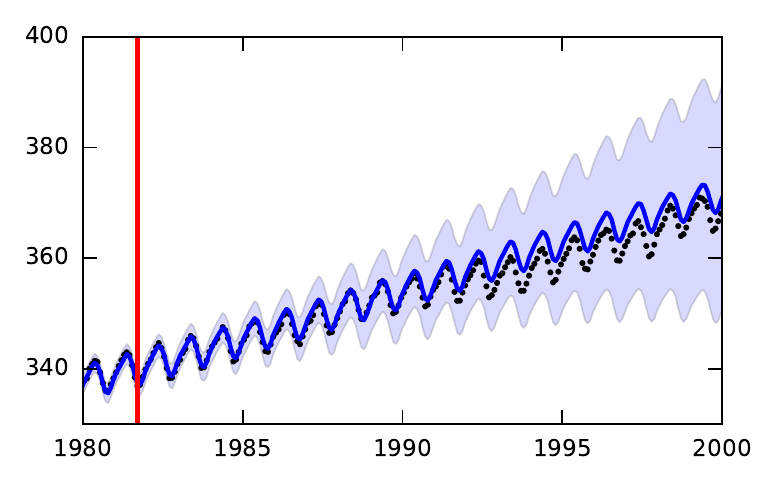}
\vspace{-1.5em}
  \caption{{\scriptsize $(\textsc{Lin} + \textsc{Per}) \times \textsc{SE}$}}
\end{subfigure}
\begin{subfigure}[b]{0.325\textwidth}
 \includegraphics[width=\textwidth]{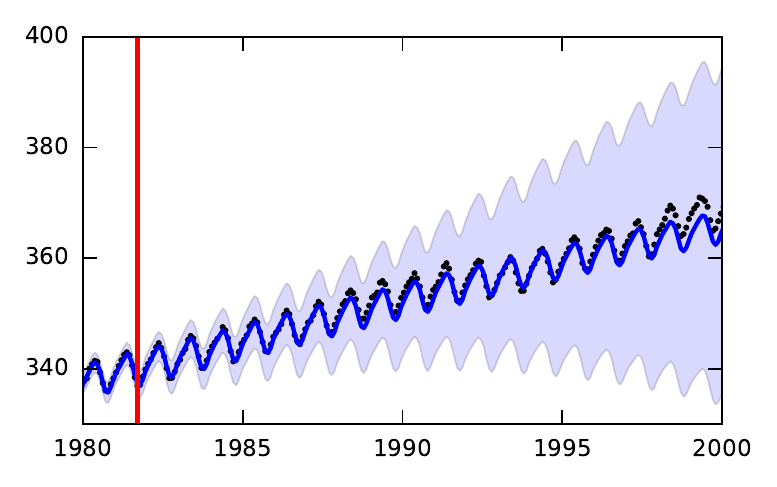}
\vspace{-1.5em}
  \caption{{\scriptsize $(\textsc{Lin} + \textsc{Per} + \textsc{RQ}) \times \textsc{SE}$}}
\end{subfigure}
\begin{subfigure}[b]{0.325\textwidth}
\includegraphics[width=\textwidth]{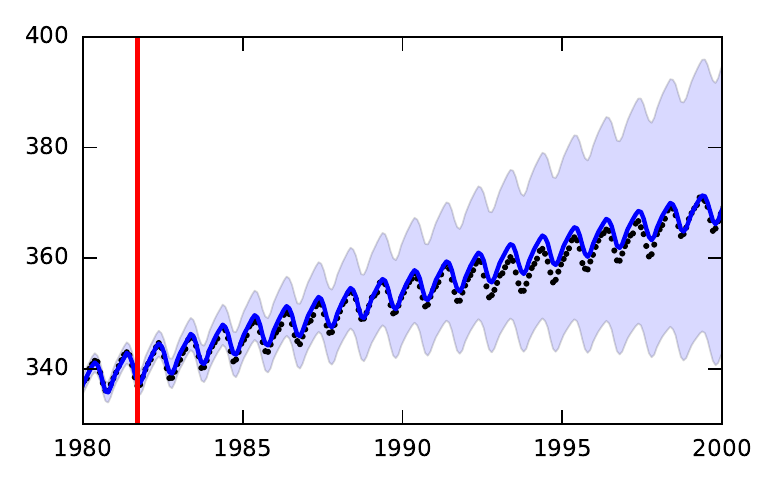}
\vspace{-1.5em}
\caption{{\scriptsize $(\textsc{Lin} + \textsc{Per}+\textsc{RQ}_1 \times \textsc{RQ}_2) \times \textsc{SE}$}}

\end{subfigure}
\caption{Three particles generated in our approach demonstrate how competing explanations, differing both in kernel structure and kernel parameters, combine to produce posterior estimates. 
	\vspace{-5pt}
	}
\label{fig:particles}
\end{figure}
\vspace{10pt}
\section{Model \& inference}
\vspace{-5pt}
Given a prior distribution over the kernel $p(k,\theta)=p(k)p(\theta|k)$, where $k$ denotes the composite kernel structure and $\theta$ the kernel parameters, and assuming a Gaussian process prior $p(f|\mathcal{D}, k, \theta)$, the posterior for predictions $y^\star$ at $x^\star$ can be written as
\begin{equation}
\label{eq:2}
p(y^\star|x^\star, \mathcal{D}) = \iint p(k, \theta) \left[\int p(y^\star|x^\star, f)p(f|\mathcal{D}, k, \theta) \diff{f}\right]  \diff{k}\diff{\theta}.
\end{equation}
Taking $p(y^\star|x^\star, f)$ to be a Gaussian likelihood, the integral over $f$ can be computed in closed form. The key task is thus to define a suitable kernel prior and perform integration over $k$ and $\theta$.  Using a sampling based system for this integration produces a weighted mixture of GPs posterior, retaining an analytic predictive distribution.

We define $p(k)$ using a probabilistic context free grammar \cite{manning1999foundations} with production rules $R = \{s\to s+s|s\times s|b\in\mathcal{B}\}$, where $s$ is a non-terminal symbol and $b$ is a base kernel, and fixed context free rule probabilities $\{p_r:r\in R\}$. We use a separable prior on $\theta$, $p(\theta|k) = \prod_{i=1}^{P}p(\theta_i|k)$, with each term determined by type of the hyperparameter $\theta_i$ and its corresponding base kernel in $k$. Jointly, these define a generative model for $k_\theta$ that we can sample from.

We marginalise over $k$ and $\theta$ using a population Monte Carlo method \cite{cappe2012population}
inspired by the use of Monte Carlo methods in decision tree inference \cite{lakshminarayanan2013top, chipman1998bayesian}.
We sample initial particles $(k^j,\theta^j)$ for $j=1,\dots,M$
from the joint prior and repeatedly mutate them using a `resample-move' algorithm \cite{gilks2001following}. 
At each iteration, we propose a mutation to each particle, using both
\begin{enumerate}
\item changing the kernel structure by sampling an existing component base kernel $b$ from the composite kernel $k$ and replacing it using one of: $b+b'$, $b\times b'$, $b$ or $\emptyset$, where $b'$ is a new randomly chosen base kernel and $\emptyset$ indicates removing $b$ from $k$; and
\item applying multiple Hamiltonian Monte Carlo (HMC) transitions to $\theta$, given the composite kernel $k$.
\end{enumerate}
This produces a new, modified set of particles drawn from a proposal distribution which can be described as a mixture of Markov kernels; 
with appropriate weights \cite{del2006sequential}, we can perform a resampling step to construct a new particle set which again approximates the posterior.
Posterior predictive comparisons to ABCD are shown in figure \ref{fig:main_result} and example individual particles are shown in figure \ref{fig:particles}.

\section{Discussion}

We have developed a fully probabilistic approach to structure discovery in time series and implemented it in Anglican. Through capturing the uncertainty in both the structure of the kernel and the hyperparameters, the system provides better calibrated uncertainty estimates than previous greedy optimisation approaches. The system allows for the posterior distribution over models to be inspected to get an understanding of how and why the system has made its predictions;
one can think of the produced samples as providing different possible explanations for the data generation.
Thus, the system provides not only a powerful regressor in a rich class of models, but also interpretable output.

\renewcommand*{\bibfont}{\normalsize}
\setlength{\bibsep}{2pt plus 0.5ex}
\bibliographystyle{abbrvnat}
\bibliography{as_paper}

\end{document}